\documentclass[10pt,twocolumn,letterpaper]{article}
\usepackage{ctable}
\usepackage{iccv}
\usepackage{times}
\usepackage{epsfig}
\usepackage{graphicx}
\usepackage{amsmath}
\usepackage{amssymb}
\usepackage{savesym}
\savesymbol{checkmark}
\usepackage{dingbat}
\usepackage{commath}
\usepackage{color,soul}
\usepackage{setspace}

\usepackage[breaklinks=true,bookmarks=false]{hyperref}

\iccvfinalcopy % *** Uncomment this line for the final submission

\def\iccvPaperID{10969} % *** Enter the ICCV Paper ID here

\ificcvfinal\fi

\begin{document}

%%%%%%%%% TITLE
\title{TriPose: A Weakly-Supervised 3D Human Pose Estimation via Triangulation from Video}
% \title{ TriPose:Triangulation from Video for Weakly-Supervised Human Pose Learning}

\author{Mohsen Gholami \and Ahmad Rezaei \and Helge Rhodin \and Rabab Ward \and  Z. Jane Wang \\
University of British Columbia\\
{\tt\small \{mgholami, rababw, zjanew\}@ece.ubc.ca}  {  \tt\small ahnr@mail.ubc.ca}  {\tt\small    rhodin@cs.ubc.ca}}
% \and

\maketitle
\ificcvfinal\thispagestyle{empty}\fi

%%%%%%%%% ABSTRACT
\begin{abstract}
Estimating 3D human poses from video is a challenging problem. The lack of 3D human pose annotations is a major obstacle for supervised training and for generalization to unseen datasets. In this work, we address this problem by proposing a weakly-supervised training scheme that does not require 3D annotations or calibrated cameras. The proposed method relies on temporal information and triangulation. Using 2D poses from multiple views as the input, we first estimate the relative camera orientations and then generate 3D poses via triangulation. The triangulation is only applied to the views with high 2D human joint confidence. The generated 3D poses are then used to train a recurrent lifting network (RLN) that estimates 3D poses from 2D poses. We further apply a multi-view re-projection loss to the estimated 3D poses and enforce the 3D poses estimated from multi-views to be consistent. Therefore, our method relaxes the constraints in practice, only multi-view videos are required for training, and  is thus  convenient for in-the-wild settings. At inference, RLN merely requires single-view videos. The proposed method outperforms previous works on two challenging datasets, Human3.6M and MPI-INF-3DHP. Codes and pretrained models will be publicly available.

\end{abstract}

%%%%%%%%% BODY TEXT
\section{Introduction}
Monocular 3D human pose estimation has gained much attention due to its various applications. It is however an ill-posed problem and most of the proposed solutions rely on supervised training that requires 3D annotations \cite{Martinez_2017_ICCV, pavlakos2018learning, Hossain_2018_ECCV, Sun_2018_ECCV, Chen_2017_CVPR, Kolotouros_2019_ICCV, Pavlakos_2018_CVPR, Pavlakos_2017_CVPR}. Obtaining 3D human annotations is not trivial and the available datasets are mostly specific to the lab settings. Therefore, generalization to in-the-wild applications remains challenging. Weakly-supervised methods were proposed to address this problem using unpaired 2D and 3D annotations \cite{Wandt_2019_CVPR, Wang_2019_ICCV, Kundu_2020_CVPR}, limited available 3D annotations \cite{Rhodin_2018_CVPR, Pavllo_2019_CVPR}, or calibrated multi-view recordings \cite{rhodin2018unsupervised, Rhodin_2018_CVPR}. However, obtaining such information for the unsupervised learning task is still an obstacle. To the best of our knowledge, there are only a few works that propose weakly-supervised training schemes without using any 3D annotation \cite{Kocabas_2019_CVPR, Iqbal_2020_CVPR, wandt2020canonpose, Wang_2019_ICCV}. \cite{Iqbal_2020_CVPR} and \cite{wandt2020canonpose} propose multi-view consistency as a supervision while \cite{Kocabas_2019_CVPR} generates pseudo ground-truth 3D poses using epipolar geometry. Although multi-view consistency is a promising self-supervision element, it is relatively a weak cue. %It is possible that 3D poses from different views which are enforced to be consistent are still erroneous. 
3D poses from different views that are enforced to be consistent may still be erroneous. Generating pseudo ground truth \cite{Kocabas_2019_CVPR} by triangulation is another approach that has shown promising performance. However, the final error is limited to the error of the generated pseudo ground truth. 
\begin{figure}[t]
\begin{center}
% \fbox{\rule{0pt}{2in} \rule{.4\linewidth}{0pt}}
\includegraphics[scale=0.25]{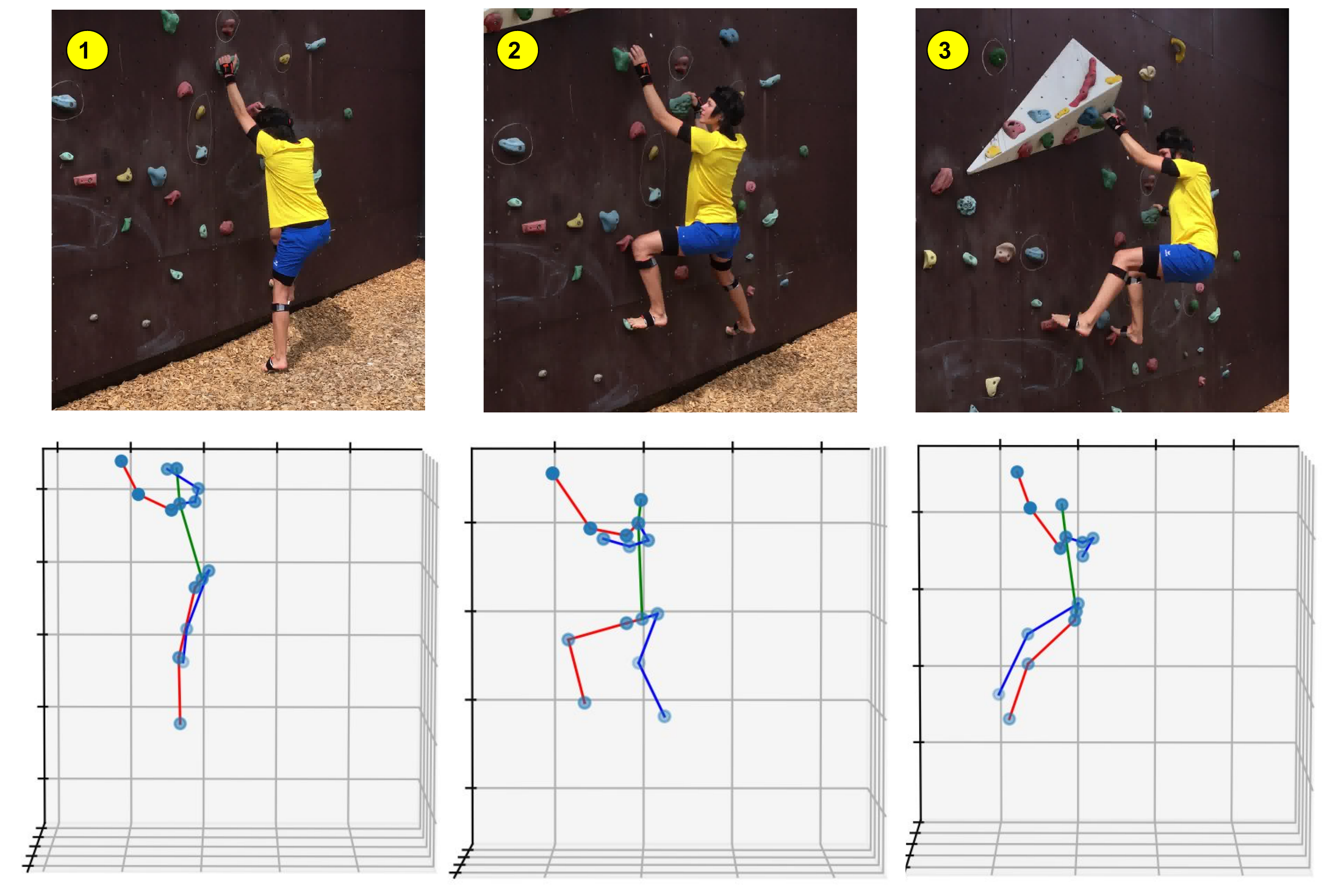}
\end{center}
  \caption{Sample predictions of the proposed weakly-supervised method for in-the-wild videos from 3DPW dataset. We note that human joints that are occluded in some frames can be estimated correctly.}
\label{fig:tiser}
\end{figure}

\begin{figure*}[t]
\begin{center}
% \fbox{\rule{0pt}{2in} \rule{.4\linewidth}{0pt}}
\includegraphics[scale=0.59]{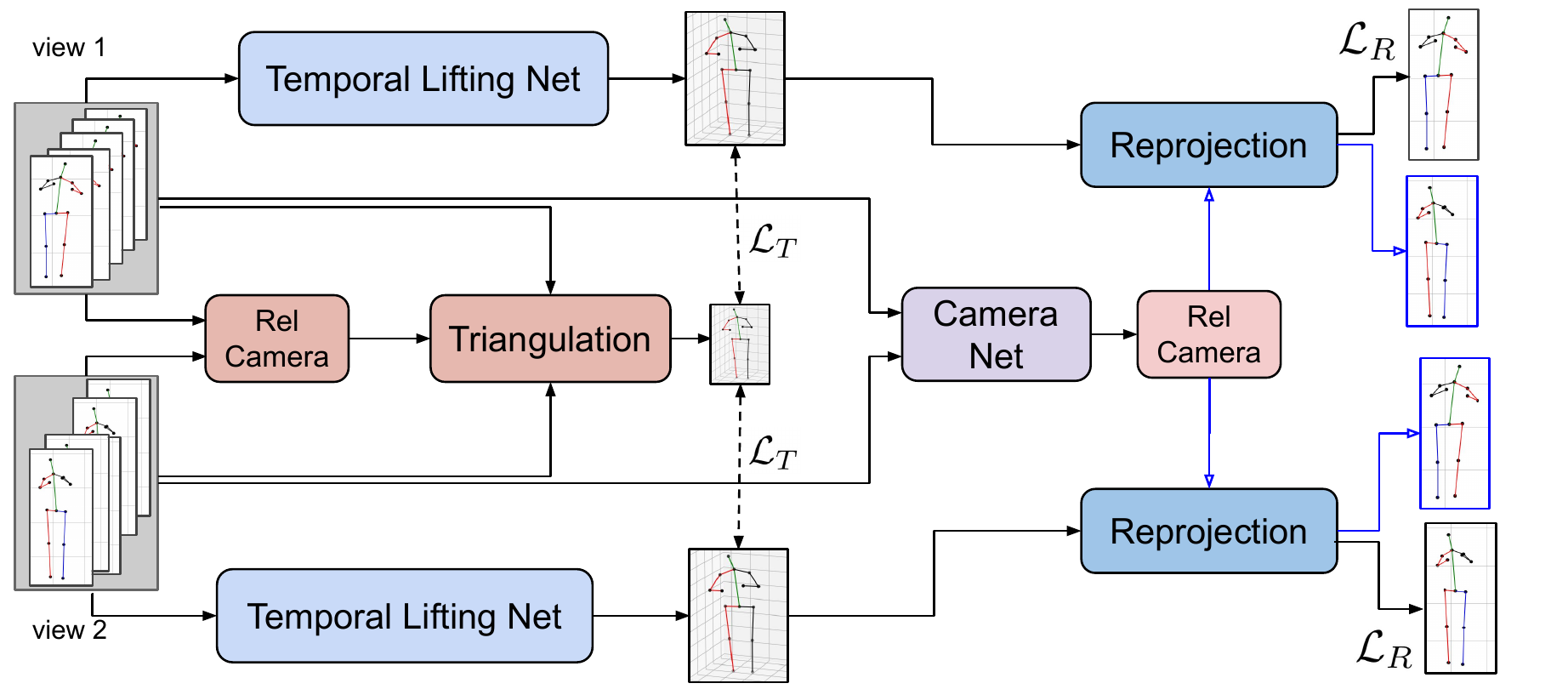}
\end{center}
  \caption{Overview of the proposed weakly supervised pose estimation approach. Video frames from two views are firstly used to estimate the relative camera parameters and also to triangulate a 3D pose in the cameras coordinate system. The triangulated 3d poses are used for training the lifting network. The 3D predictions are re-projected to multi-view 2D to further supervise the lifting network.}
\label{fig:overview}
\end{figure*}

In this work, we propose a method that combines multi-view cameras with triangulation supervision while leveraging temporal dependency in video using a temporal lifting network. This approach needs neither 3D annotations nor calibrated cameras for training. Fig. \ref{fig:tiser} illustrates the performance of our method in unseen videos, and  Fig. \ref{fig:overview} shows the overview of the proposed method. The relaxations made in this work avail the training with in-the-wild multi-view images. Inspired by EpipolarPose \cite{Kocabas_2019_CVPR}, we generate 3D poses using triangulation. However, we integrate camera joint confidence scores into the triangulation to choose sufficiently good views. This step helps diminish wrong triangulation for challenging datasets such as MPI-INF-3DHP \cite{mehta2017monocular}. In fact, instead of triangulating complete but poor pseudo ground truths for the lifting network, we choose to generate pseudo ground truths that are sufficiently good.

The generated pseudo ground truth is only taken as a guidance. We further enforce estimated 3D poses to be consistent from different views using multi-view re-projection, which essentially addresses the wrong depth estimation and tackles possible errors in triangulation.
Note that even though this re-projection loss is equivalent to the preprocessing triangulation objective, it behaves differently since it is used to train the neural network parameters that have to fit consistently to all training images as well as being temporally consistent, as we explain below.

Despite previous weakly-supervised works \cite{Kocabas_2019_CVPR, Iqbal_2020_CVPR, wandt2020canonpose}, we propose learning from video instead of a single frame. Pavllo et al. \cite{Pavllo_2019_CVPR} use a temporal convolution network for semi-supervised training using re-projection loss. However, their re-projection scheme requires the root trajectory to be estimated. Estimating root trajectory from video is a difficult task. Therefore they require a small set of 3D annotations to initialize the training. Kocabas et al. \cite{Kocabas_2020_CVPR} use a temporal motion discriminator in an adversarial training scheme, which uses an archive of ground truth 3D poses. Despite the limitations, both \cite{Pavllo_2019_CVPR} and \cite{Kocabas_2020_CVPR} demonstrate the advantage of exploiting the temporal information in the video. The temporal structure helps with the problem of occluded body parts in a single frame and reduces jittery motions, as the network can benefit from the pose information of previous frames. Similarly, we employ a recurrent-based temporal lifting network to map 2D poses to 3D ones. By contrast to previous approaches, we do not require any 3D annotation for training.

Single-view re-projection from 3D to 2D has been used in previous weakly-supervised works \cite{Kanazawa_2018_CVPR, Wandt_2019_CVPR}. Due to perspective ambiguity, a single 2D pose corresponds to many different 3D poses. Therefore, re-projection alone is insufficient for training and should be accompanied by other supervision elements such as unpaired 3D annotations \cite{Wandt_2019_CVPR} or limited 3D annotations \cite{Pavllo_2019_CVPR}. In this work we use multi-view re-projection, which is often more convenient to obtain than unpaired/limited 3D annotations. Multi-view re-projection does not suffer from perspective ambiguity since there is only one 3D pose that can be accurately re-projected to multi-view 2D planes.

We evaluate our method on two public datasets, namely Human3.6M \cite{h36m_pami} and MPI-INF-3DHP \cite{mehta2017monocular}, and achieve state-of-the-art results. We also qualitatively evaluate the performance of our method on in-the-wild videos. The contributions of our work are as follows:
\begin{spacing}{0.5}
\begin{itemize} 
\item  Proposing a weakly supervised training scheme that requires neither camera extrinsic nor any 3D annotations.
\item  Proposing a temporal lifting network that is tailored for and supports self-supervised training.
\item  Proposing a multi-view re-projection method that enforces view-consistency.
\item  Providing in-depth comparisons with existing supervision strategies and state-of-the-art results on two most widely used 3D pose estimation benchmarks.
\end{itemize}
\end{spacing}

\section{Related Work}
\textbf{Full-3D Supervision.}
Fully supervised 3D human pose estimation methods can be broadly divided into two categories: i) end-to-end training that accepts image or video as input and directly estimates 3D poses \cite{Tome_2017_CVPR, 10.1007/978-3-319-49409-8_15, Xu_2020_CVPR, Sun_2018_ECCV,Li_2020_CVPR}; ii) two-step methods that firstly estimate 2D skeleton and then lift this 2D pose to 3D space \cite{Chen_2017_CVPR, mehta2020xnect, Pavlakos_2018_CVPR,Pavllo_2019_CVPR, Tekin_2017_ICCV}. The second approach has the advantage of intermediate supervision from large in-the-wild 2D annotations. Recent works have shown promising results using lifting methods \cite{Martinez_2017_ICCV, Pavllo_2019_CVPR} and our approach falls in this category. 

\textbf{Temporal Supervision.}
Temporal information provides further knowledge of previous and future human motions, which can help with occlusion and jittery estimations \cite{Arnab_2019_CVPR, Pavllo_2019_CVPR, Cheng_2019_ICCV, Hossain_2018_ECCV, Kocabas_2020_CVPR, Cheng_2019_ICCV, Kanazawa_2019_CVPR}. A long short-term memory (LSTM) sequence to sequence method was proposed to exploit temporal information in video and predict a sequence of 3D poses from a sequence of 2D poses \cite{Hossain_2018_ECCV}. Sequence to sequence pose models were shown to be prone to drift \cite{Pavllo_2019_CVPR}. Pavllo et al. \cite{Pavllo_2019_CVPR} propose a dilated convolutional network to explore long history information, while being robust to drift in a long sequences. \cite{Kocabas_2020_CVPR} also uses a gated recurrent unit (GRU) encoder to encode the sequence of pose information generated by convolutional layers. Our approach is close to \cite{Kocabas_2020_CVPR}, however we utilize a residual network \cite{Martinez_2017_ICCV} to decode pose information. We show that a many-input to single-output recurrent lifting network achieves better results than that of the sequence to sequence model \cite{Hossain_2018_ECCV}. 

\textbf{Unpaired/Limited 3D supervision.}
Obtaining limited or unpaired 3D annotations is more convenient than full 3D annotations. Employing generative adversarial networks have been extensively investigated to use unpaired 2D and 3D data \cite{Kanazawa_2018_CVPR, Wandt_2019_CVPR, Yang_2018_CVPR, Kundu_2020_CVPR, Chen_2019_CVPR}. In these works, the generator is expected to estimate 3D fake poses that are as plausible as the real poses so that the discriminator can not distinguish between real and fake 3D poses. The re-projection error has been commonly used along with adversarial losses to further supervise the training \cite{Wandt_2019_CVPR, Kanazawa_2018_CVPR}. Partial 3D annotation has also been showed to be effective for semi-supervised training along with re-projection  \cite{Pavllo_2019_CVPR, Cheng_2019_ICCV} or multi-view consistency loss \cite{Rhodin_2018_CVPR,Mitra_2020_CVPR}. The idea of multi-view consistency is that the estimated 3D poses from different views should be the same. The multi-view constraint has also been used to learn geometry of human poses in \cite{Chen_2019_CVPR, rhodin2018unsupervised}, which lessens the need for 3D annotations. 

\textbf{Supervision without 3D.}
Recently, there is more interest in training schemes that are fully independent of 3D annotations. The required additional supervision can be provided by cycle consistency loss \cite{Drover_2018_ECCV_Workshops,Chen_2019_CVPR} or multi-view data \cite{Iqbal_2020_CVPR, Wandt_2019_CVPR, Kocabas_2019_CVPR}. This relaxation facilitates training for many in-the-wild applications like field sports (hockey, football) \cite{Cai_2019_CVPR_Workshops} that multi-view data is already available. To our knowledge, there are only three similar works in the literature that merely need multi-view data as input \cite{Kocabas_2019_CVPR,Iqbal_2020_CVPR,wandt2020canonpose}. Kocabas et al. \cite{Kocabas_2019_CVPR} generate pseudo 3D ground truth by triangulation from multi-view data. This simple approach was shown to be effective if 2D keypoints are sufficiently accurate. The main limitation of this approach is the lack of robustness to error in the input 2D keypoints. Iqbal et al. \cite{Iqbal_2020_CVPR} propose a 2.5 training method that disentangles depth from x and y coordinates. They use an independent 2D annotation for estimating x and y. For depth estimation, assuming that the estimated 3D pose from all views should be the same, they use a multi-view consistency loss between estimated 3D poses from different views. 
Although multi-view consistency has been shown to be strong, the indirect supervision remains vague when two wrong 3D poses are consistent. \cite{wandt2020canonpose} proposes a multi-view re-projection of estimated 3D poses back to 2D. 
Instead of simply using a loss between multi-view estimated 3D poses, \cite{wandt2020canonpose} defines the loss between re-projected 2D poses. Comparing with multi-view consistency in 3D, multi-view re-projection to 2D has been shown to be more promising for view-consistency enforcing. 

In the absence of 3D annotations, accurate 2D poses are important for accurate multi-view re-projection. Occluded body parts can lean to degenerated solutions. Therefore, we encode information from a sequence of 2D keypoints using a temporal lifting network. Similar to EpipolarPose \cite{Kocabas_2019_CVPR}, we generate pseudo ground truth 3D annotations to train the lifting network. However, to ensure triangulation is only done from appropriate views, we also incorporate confidence of 2D joint. Furthermore, we employ a multi-view re-projection loss so that the training is not fully dependent on pseudo ground truth annotations. The overall method is shown in Fig. \ref{fig:overview}.    

% \subcetion{Semi-supervised Methods.}
% \subsection{Weakly Supervised Methods.}
\section{Method}
Our framework requires un-calibrated multi-view videos for training and can be applied to a single video at inference time. During training, we use triangulation to produce a rough pseudo ground truth 3D annotation and self-calibrated camera orientation. We then train a recurrent lifting network to estimate the single 3D pose from a sequence of input 2D poses. The training criteria are the least-square loss to the pseudo 3D annotations and the multi-view re-projection loss. With this design, inference only requires single-view videos. 

\subsection{Triangulation}
\label{sec:triangulation}

We consider a set of 2D keypoints from two views as inputs, $X_{1,j}$, $X_{2,j}$, where $j$ is in range of 2D keypoints. We also accept joint confidences $[c_{1,1},...,c_{1,J}]$, $[c_{2,1},...,c_{2,J}]$ from two views where $J$ is the number of human joints. If the average confidence of any view is less than 0.8 or if the confidence of any of joint is less than 0.7, we exclude the view from triangulation. The numbers were selected empirically. The epipolar constrain is then applied to $X_{1,j}$, $X_{2,j}$ through
\begin{align}
X_{1,j} \mathbf {F} X_{2,j} = 0, 
\label{equ:wasserstein_loss}
\end{align}
where $\mathbf{F}$ is fundamental matrix. The above constrain can be reformulated to $\mathbf{A}f=0$, where $f$ is a vector form of $\mathbf{F}$. It is solved by singular value decomposition (SVD) to find the fundamental matrix and remove outliers with RANSAC over a set of two views. The threshold for consensus set of RANSAC was set to $3/(f_1+f_2)$ where $f_1$ and $f_2$ are focal length of two cameras. The level of confidence for estimated matrix to be correct was set to 0.999. From the fundamental matrix, we compute the relative camera orientation using essential matrix decomposition, $\mathbf{E}=\mathbf{K}^T \mathbf{F} \mathbf{K}$ 
We consider the first camera as the center of coordinate frame. Therefore, rotation matrix, $\mathbf{R}$, and transition vector, $\mathbf{t}$, of the first camera are an identical matrix and zero, respectively. The essential matrix decomposition has four solutions with different camera rotations. We use charity check to remove those camera parameters that after triangulation return a negative depth in the camera coordinate frame. 
%The charity check simply checks if the reconstructed 3D poses are in-front of the camera. 

The estimated $\mathbf{R}$ and $\mathbf{t}$ are used to reconstruct 3D poses by polynominal triangulation \cite{hartley_triangulation_1997}. The triangulated 3D poses are in the camera coordinate system of the first view. They can be transferred to the second view using $\mathbf{R}$. The output of triangulation is not scaled to actual human skeleton. Therefore, we use the average body bone length of the training set to adjust the scale of 3D poses. The triangulation loss is defined as mean per joint position error between the triangulated 3D pose $\hat{Y}$ and estimated 3D pose $Y$ 
\begin{align}
\mathcal{L}_{T} = \norm{\hat{Y}-Y}_2. %\mathcal{F}(C_1, C_2),
\label{equ:wasserstein_loss}
\end{align}
% where $C_i=[c_{i,1}, c_{i,2}, ..., c_{i,j}]$ is joint confidences of each views and $\mathcal{F}(C_1, C_2)$ is as follows:
% \begin{equation}
%   \mathcal{F}(C_1, C_2)=
%     \begin{cases}
%       0  & \mbox{if  } c_{i,j}\\
%       0  & \mbox{if  } \overline{C_i}<b\\
%       (C_1+ C_2)/2 & \text{otherwise}\\
%     \end{cases}       
% \end{equation}
% where $a$ and $b$ thresholds where empirically set to 0.7 and 0.8, respectively.

\subsection{View-Switching re-projection}
Re-projecting the estimated 3D pose to image plane should generate a 2D pose identical to the input 2D in an ideal case. However, a 2D pose corresponds to many 3D poses due to perspective ambiguity and single-view re-projection can not disambiguate wrong depth estimates. On the other hand, re-projection to multi-view enforces correct depth estimation since there is only one point in 3D space that can be exactly projected to corresponding points in multiple views. In this work we use multi-view re-projection on an estimate of the relative camera orientation between multiple views. Therefore, we do not need camera extrinsic parameters for same-view projection. Furthermore, we use a weak perspective re-projection,
\begin{gather}
 X^{(i,j)}_{rep} =
  \begin{bmatrix}  
   1 & 0 & 0 \\
   0 & 1 & 0 
   \end{bmatrix} \mathbf{R_{ij}} X_{3D},
\label{equ:projection}
\end{gather}
% Therefore, we neither require camera intrinsic parameters nor exact camera translation.
% In order to reproject to other camera views we estimate the relative rotation matrix between two views. Particularly, we estimate $\mathbf{R_{ij}}$ which is the relative rotation matrix between view  $\mathbf{i}$ and $\mathbf{j}$.\hr{[This is redundant. You explain it already in the previous section, then in the paragraph above and here again. Can you introduce the $_{ij}$ notation right away and re-use it here?]} 
$\mathbf{R_{ij}}$ is the relative rotation matrix between view  $\mathbf{i}$ and $\mathbf{j}$.
For re-projection to the same camera, $\mathbf{R_{ij}}$ is equal to the identity. %Estimated 3D poses are in camera coordinate system.
The relative camera parameter is initialized using the RANSAC algorithm in the previous step. We explain in Section~\ref{sec:camera net} how this estimate  is further refined with a lifting network. Since we are not estimating the root joint trajectory, the re-projected 2D poses are all in the same scale and there is an offset between root joints. Therefore, we perform a root centering and then normalization to the reconstructed and input 2D. The re-projection loss over all camera pairs $i,j$ is as follows

\begin{align}
\mathcal{L}_{R} = \sum_{i=1}^{n}\sum_{j=1}^{n} \norm{\frac{X_{2D}}{\norm{X_{2D}}}-\frac{X^{(ij)}_{rep}}{\norm{X^{(ij)}_{rep}}}},
\label{equ:projection_loss}
\end{align}
where $X_{2D}$ and $X^{(ij)}_{rep}$ are actual and re-projected 2D skeletons and $n$ is number of views.

% Since we have relative rotation and transition matrix we can perform reproject to all available views following below formula  

% \begin{gather}
%  \begin{bmatrix} x_p \\ y_p \\ 1   \end{bmatrix} =
%   \begin{bmatrix}  
%   f_x & 0 & c_x \\
%   0 & f_y & c_y \\
%   0 & 0 & 1 
%   \end{bmatrix} \begin{bmatrix} \frac{x_c}{z_c} \\ \frac{y_c}{z_c}\\ 1  \end{bmatrix},
%   \begin{bmatrix} x_c \\ y_c \\ z_c   \end{bmatrix}
%  = [R|T]  \begin{bmatrix} x_w \\ y_w \\ z_w \\ 1 \end{bmatrix}
% \end{gather}

\subsection{Lifting Network}
\begin{figure}[t]
\begin{center}
% \fbox{\rule{0pt}{2in} \rule{.9\linewidth}{0pt}}
\includegraphics[scale=0.8]{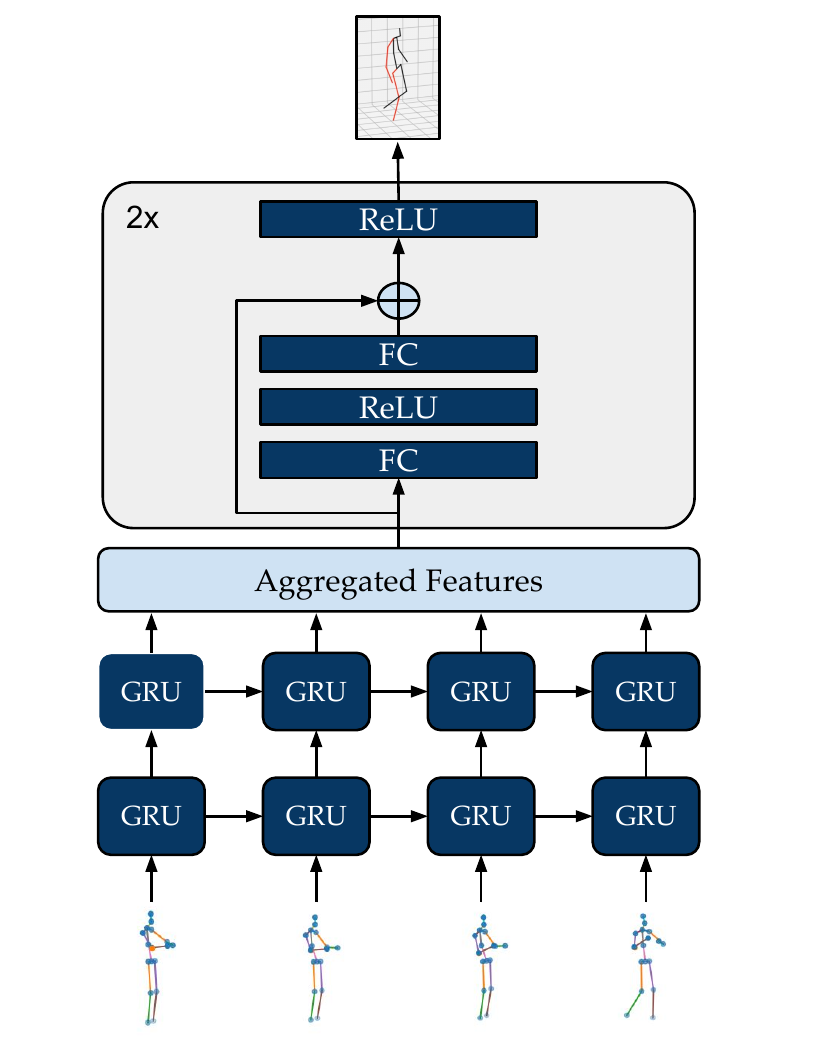}
\end{center}
  \caption{Overview of our temporal lifting network.}
\label{fig:liftingnet}
\end{figure}

Figure \ref{fig:liftingnet} illustrates structure of our lifting network. We encode information from previous and future time frames using two layers of GRUs with size of 1024. Considering an input sequence of $[X_{t-T},...,X_{t+T}]$, GRU layers return the hidden state of sequence as $[h_{t-T},...,h_{t+T}]$. Instead of using the final hidden state, we perform a maxpooling and average pooling over the sequence of hidden states and concatenate the two representations. The learnt information from GRU layers are then fed to blocks of residual neural network. The second part of the network has 2 fully connected residual blocks, each block with two fully connected layers with 1000 neurons. Kocabas et al. \cite{Kocabas_2020_CVPR} use two layers of GRU with self-attention mechanism as a motion discriminator. We use GRU layers with static pooling layers that aggregates information from all hidden layers. \cite{Kocabas_2020_CVPR} shows that using self-attention mechanism after GRU layers helped to dynamically learn important time frames. However, we did not observe a difference between static feature aggregation and self-attention mechanism in our experiments. \cite{Hossain_2018_ECCV} uses the information from the last hidden layer of an LSTM-based encoder and passes it to a decoder. We aggregate sequence of hidden layers and then use two fully connected residual blocks to return 3D poses following the work in \cite{Martinez_2017_ICCV}. We do not use batch normalization and dropout of \cite{Martinez_2017_ICCV}. 

For training the lifting network, two loss functions are used. This includes 2D multi-view re-projection loss ($\mathcal{L}_{R}$) and triangulation loss ($\mathcal{L}_{T}$). The total lifting network loss is as follows
\begin{equation}
\mathcal{L}=\mathcal{L}_{R}+\mathcal{L}_{T}.
\label{equ:lifting_loss}
\end{equation}

\subsection{Camera Correction Network}
\label{sec:camera net}

We use the same structure of two GRU layers and two residual blocks for a camera network to estimate the relative rotation matrix $\mathbf{R}\in \mathbb{R}^{3\times3}$ between the two input views. Despite the lifting network that has only one sequence of 2D poses as input, the input to the camera network consists of two sequences of 2D poses from both views. This input allows the network to learn the relative rotation matrix of two input views. Despite previous works that estimate a rotation matrix for each camera \cite{wandt2020canonpose}, estimating the relative rotation matrix has the benefit of using multiple views as input to the camera model. Moreover, since we do not need the camera network during inference, our framework can work with single-view input in the inference time. Considering two input views, the first and second view rotation matrices are $\mathbf{R_{i}}$ and $\mathbf{R_{j}}$. The relative rotation matrix of these two views is $\mathbf{R_{ij}}=\mathbf{R_{j}}\times\mathbf{R_{i}^T}$. Camera correction network estimates  $\mathbf{\bar{R}_{ij}}$ which is defined by $\mathbf{\bar{R}_{ij}}=\mathbf{R_{ij}}-\mathbf{R_{ij}}_{(triang)}$ where $\mathbf{R_{ij}}$ is the actual rotation matrix and $\mathbf{R_{ij}}_{(triang)}$ is the rotation matrix computed in triangulation. This means that our camera network \textit{corrects} the previous rotation matrix. We optimize the parameters of camera correction network to minimize the objective function in equation \ref{equ:projection_loss}. The computed loss in the equation \ref{equ:projection_loss} back-propagates through the camera correction network as well as the lifting network.   

% in triangulation by    \hr{Not clear what losses you use? Do you optimize eq. (4) and (3) with respect to R instead of X? Be specific!}
% \hr{You also need to explain how this updated R is used, instead of the one computed by triangulation.}

\subsection{Adversarial Training}

We further analyse the performance of our lifting network for unpaired 2D and 3D annotation instead of multi-view data, as in \cite{Wandt_2019_CVPR}. In these experiments, we aim to merely evaluate performance of our temporal lifting network for common weakly supervised training schemes. The input for these experiments is a single-view video, however we have a collection of unpaired 3D human poses. This enables combining outdoor images with 2D labels and 3D poses captured in an indoor motion capture studio. We implement a Wasserstein GAN \cite{pmlr-v70-arjovsky17a} that includes a generator and a critic network. The introduced lifting network in the previous section is used as a generator. The critic network tries to maximize the distance between the output of the lifting network and batches of real 3D human pose. We use a motion critic network that accepts single 3D pose as well as a sequence of 3D poses. This motion critic is compromised of 2 layers of GRU units followed by a linear layer. The parameters of the critic are clipped between $[-0.01, 0.01]$. We also performed experiments on different critic models introduced in \cite{Wandt_2019_CVPR} and a convolutional network with four convolutional layers and a maxpooling layer in the middle. Our experiment showed that the GRU critic outperformed the others even when we do not have a sequence of 3D poses as output. We use a single-view re-projection loss, $\mathcal{L}_{R}$, and an adversarial loss, $\mathcal{L}_{adv\mathcal{G}}$, from unpaired 3D annotations for training the lifting network as follows 
\begin{equation}
\mathcal{L}=\mathcal{L}_{R}+\mathcal{L}_{adv\mathcal{G}},
\label{equ:generator_loss}
\end{equation}
where the loss functions used in adversarial training of the critic and the generator networks are defined as follows

\begin{align}
\mathcal{L}_{adv\mathcal{C}} &= \mathbb{E}_{W \sim P_R}[(\mathcal{C}(W)] - \mathbb{E}_{W \sim P_G}[(\mathcal{C}(W)], \\
\mathcal{L}_{adv\mathcal{G}} &= \mathbb{E}_{W \sim P_G}[(\mathcal{C}(W)], 
\label{equ:wasserstein_loss}
\end{align}
where $\mathcal{G}$ and $\mathcal{C}$ are generator and critic networks, respectively and $\mathcal{L}_{adv\mathcal{C}}$ is the adversarial loss for the critic network. $_{P_R}$ and $_{P_G}$ are distribution of real and generated 3D poses, respectively.

\section{Experiments}

In this section, we introduce the datasets, metrics, and the training procedure. We then compare our results with state-of-the-art self-supervision methods and perform ablation studies on different parts of our framework. We use an AlphaPose network \cite{xiu2018poseflow, fang2017rmpe} pretrained on MPII dataset \cite{Andriluka_2014_CVPR} as the 2D backbone. The input 2D poses are normalized using frame height $(h)$ and width $(w)$ so that $[0, w]$ is mapped to $[-1, 1]$ while preserving the aspect ratio. Following previous works \cite{Martinez_2017_ICCV,Pavllo_2019_CVPR,Hossain_2018_ECCV}, we estimate the 3D poses in camera coordinate systems and all 3D joints are relative to the root joint (pelvis). We follow \cite{Kocabas_2019_CVPR} and freeze the triangulation part during training. The learning rate was 0.001 and Adam optimizer was used in all of the experiments except for the critic network. A stochastic gradient descent optimizer was used for the critic network.    
\subsection{Dataset}
We use three different datasets to perform analysis on our framework. We also use in-the-wild videos to further evaluate plausibility of the predictions.

\textbf{Human3.6M} \cite{h36m_pami} is the most popular 3D human dataset that includes videos from 4 views recorded in an indoor setting. The dataset is from 7 subjects that perform 15 challenging actions. We use the standard splitting of training and test sets, and use subjects 1,5,6,7,8 for training and subjects 9 and 11 for testing. We use the original 50Hz frame rate without down-sampling the video. We do not perform data augmentation on the dataset.

\textbf{MPI-INF-3DHP} \cite{mehta2017monocular} has been recorded in a lab setting, same as Human3.6m dataset. However, the test set has outdoor recorded videos. The dataset includes 14 camera views. We follow the standard protocol and use 5 chest camera views for training. We use the original frame rate of the dataset for training and testing. 

\textbf{3DPW} \cite{vonMarcard2018} has been recorded fully in-the-wild. The dataset includes videos from a single view. The camera is moving while the subject performs different daily activities in street, park, or indoors. We use this dataset for qualitative analysis. We also use a sample ski video from YouTube for further evaluation. 

% \textbf{MPII} \cite{andriluka14cvpr} provides 2D annotations that includes 16 keypoints. The 2D estimator used in this study has been pretrained on MPII.

\subsection{Metrics}
We use three variants of the mean per joint position error (MPJPE). It computes the average euclidean distance between true and estimated 3D poses. Since our method is self-supervised, the scale of estimated 3D poses can be arbitrary. We therefore normalize triangulated poses by the average bone lengths in the training set as explained in Section~\ref{sec:triangulation}. 
Following previous work, we also report the NMPJPE, the MPJPE after normalizing scale, which does not require using the training set lengths.
In addition, we perform Procrustes alignment on the estimated 3D poses and compute the MPJPE on the aligned pose (PMPJPE). PMPJPE thereby excludes rotation, scale, and transition errors.

{\renewcommand{\arraystretch}{1}
\begin{table}[t]
\footnotesize
\centering
\caption{Estimation results of different networks for the Human3.6M dataset. All values are in millimeters and the lower the better. UP-3D means unpaired 3D and MV means multi-view. The check-mark in the 2D column indicates that the ground truth 2D of training set is used. The best results are marked bold.}
\label{tab:Human3.6M}
\begin{tabular}{ p{1.8cm} p{0.1cm} p{1.3cm} c c c  }

\hline
Method & 2D & Supervision &  MPJPE & NMPJPE & PMPJPE \\
\hline 
Martinez \cite{Martinez_2017_ICCV}
& \checkmark & 3D & 62.9 & - & 47.7  \\
Hossain \cite{Hossain_2018_ECCV}
& \checkmark & 3D & 65.9 & - & 41.5 \\
Hossain \cite{Hossain_2018_ECCV}
& \checkmark & 3D & 58.3 & - & 44.1 \\
Iqbal \cite{Iqbal_2020_CVPR}
& \checkmark & 3D& 55.5 &  51.4 &  41.5 \\
Pavllo \cite{Pavllo_2019_CVPR}
& \checkmark &3D & 46.8 &  - &  36.5 \\
Cheng \cite{Cheng_2019_ICCV}
& \checkmark &3D& \textbf{46.7} & - & \textbf{34.1}\\
Ours-baseline&\checkmark &3D & 55.4 & 52.0 & 41.7 \\
\hline
Wandt \cite{Wandt_2019_CVPR} 
& \checkmark&UP-3D & 89.9 &  - &  66.1 \\

Kanzawa \cite{Kanazawa_2018_CVPR} 
& \checkmark&UP-3D & 169.5 &  - &  66.5 \\
Kolotouros \cite{Kolotouros_2019_ICCV}  
& \checkmark&UP-3D & - &  - &  62.0 \\
Kundu \cite{Kundu_2020_CVPR}
& \checkmark&UP-3D & 85.8 &  - &  - \\
\hline
% Kocabas \cite{Kocabas_2019_CVPR}
% & \checkmark & MV+R &55.08 &  54.90 &  47.91 \\
% Ours  
% & \checkmark & MV+R & \textbf{} & \textbf{ } & \textbf{}  \\ 
% \hline
Wandt \cite{wandt2020canonpose}
& \checkmark & MV &65.9 & - &  - \\
Iqbal \cite{Iqbal_2020_CVPR}
& \checkmark & MV&59.3 &  - &  - \\
Ours  
& \checkmark & MV & \textbf{56.7} & \textbf{54.1} & \textbf{43.8}  \\ 
\hline
Kocabas \cite{Kocabas_2019_CVPR}
& &MV & - &  77.75 &  70.67 \\
Wandt \cite{wandt2020canonpose}
& &MV & 74.3 & - &  53.0 \\
Iqbal \cite{Iqbal_2020_CVPR}
& &MV &69.1 &  66.3 &  55.9 \\
% Ours(SH 2D) & MV & - & - & - \\
Ours + GRUA
& & MV &  \textbf{64.4} & \textbf{60.9} & \textbf{48.1} \\ Ours + Pavllo
& & MV &  \textbf{62.9} & \textbf{59.5} & \textbf{47.0} \\
\hline
\end{tabular}
\end{table}}

\subsection{Quantitative Results}
Table \ref{tab:Human3.6M} shows comparison results on the Human3.6M dataset. We consider three different levels of supervision. 
First, \textit{3D}, the fully supervised baseline that uses 2D and 3D annotations of training. In this scenario, the 2D estimator has been fine-tuned on 2D annotation of the training set. 
Second, \textit{UP-MV}, in this scenario multi-view data is available for self-supervision and 2D annotations of training set are also provided for fine tuning. We obtain an error of 56.7 mm in \textit{UP-MV} which is lower than fully supervised training of recurrent temporal models in \cite{Hossain_2018_ECCV} and \cite{Kocabas_2020_CVPR}. This shows the superiority of our lifting network comparing with the previous recurrent temporal models. Our fully-supervised training error compares well with state-of-the art methods ae well as with fully-supervised training of \cite{Iqbal_2020_CVPR}. 
In the third scenario, \textit{MV}, we use only multi-view data without 2D annotation. We obtain an error of 64.4 mm using temporal GRU lifting network (GRUA) and an error of 62.9 mm using  temporal lifting network introduced by Pavllo et al. \cite{Pavllo_2019_CVPR}, which is 6 mm $(8\%)$ better than the previous state-of-the-art method \cite{Iqbal_2020_CVPR}.  \cite{wandt2020canonpose}. The gap between our weakly-supervised and fully-supervised results is only 8 mm which is smaller than the same gap from Iqbal et al.\cite{Iqbal_2020_CVPR}. There is a great margin between our results and the results from Kocabas et al \cite{Kocabas_2019_CVPR} who also use triangulation. In the next section we validate our results without temporal model and multi-view re-projection by results from \cite{Kocabas_2019_CVPR} and show that the improvements are gained from these two supervision elements.

We further evaluate the performance of our framework on 3DHP dataset with two different levels of supervision. The evaluation set of 3DHP is more challenging than Human3.6M and contains outdoor videos. Table \ref{tab:3DHP} shows our results on this dataset. In fully supervised training, our error is 9 mm less than the previous works. Our weakly-supervised result has a great margin from previous works. When compared with other weakly-supervised methods that rely on unpaired 3D annotations, we obtain comparable results with these works \cite{Kundu_2020_CVPR}, however our training scheme is more convenient since it is independent of 3D annotation.

{\renewcommand{\arraystretch}{1}
\begin{table}[t]
\footnotesize
\centering
\caption{Estimation results of different networks for the MPI-INF-3DHP dataset. All values are in millimeters and the lower the better. (*) indicates the use of extra in-the-wild data for training and (**) indicates the use of extra camera views.}
\label{tab:3DHP}
\begin{tabular}{ p{1.8cm}p{0.1cm} p{1.2cm} c c c }
\hline
Method&2D&Supervision&MPJPE&NMPJPE&PMPJPE \\
\hline
Rhodin \cite{Rhodin_2018_CVPR} 
& \checkmark &3D &-&101.5&-\\
Kokabas \cite{Kocabas_2019_CVPR}
& \checkmark &3D & 109.0 & 106.4 & - \\
Iqbal \cite{Iqbal_2020_CVPR}
& \checkmark &3D & 110.8 & 98.9 & - \\
ours & \checkmark &3D &\textbf{101.5} & \textbf{97.5} & \textbf{76.5} \\

\hline
Kanzawa \cite{Kanazawa_2018_CVPR} 
& \checkmark &UP-3D & 169.5 &  - &  113.2 \\
Kolotouros \cite{Kolotouros_2019_ICCV} 
& \checkmark &UP-3D & 124.8 &  - &  80.4 \\
Kundu (*) \cite{Kundu_2020_CVPR}
& \checkmark &UP-3D & 103.8 &  - &  - \\
\hline
Kocabas \cite{Kocabas_2019_CVPR}
& &MV+R & 126.79 &  125.65 &  - \\
Iqbal(*) \cite{Iqbal_2020_CVPR}
& &MV & 122.4 & 110.1&  \\
Wandt (**) \cite{wandt2020canonpose}
& &MV & - & 104.0 &  \textbf{70.3} \\
% Ours(SH 2D) & MV & - & - & - \\
Ours 
& &MV &  \textbf{105.6} & \textbf{101.1} & 78.5 \\ 
\hline
\end{tabular}
\end{table}}

\subsection{Qualitative Results}
\begin{figure}[t]
\begin{center}
% \fbox{\rule{0pt}{2in} \rule{.4\linewidth}{0pt}}
\includegraphics[scale=0.23]{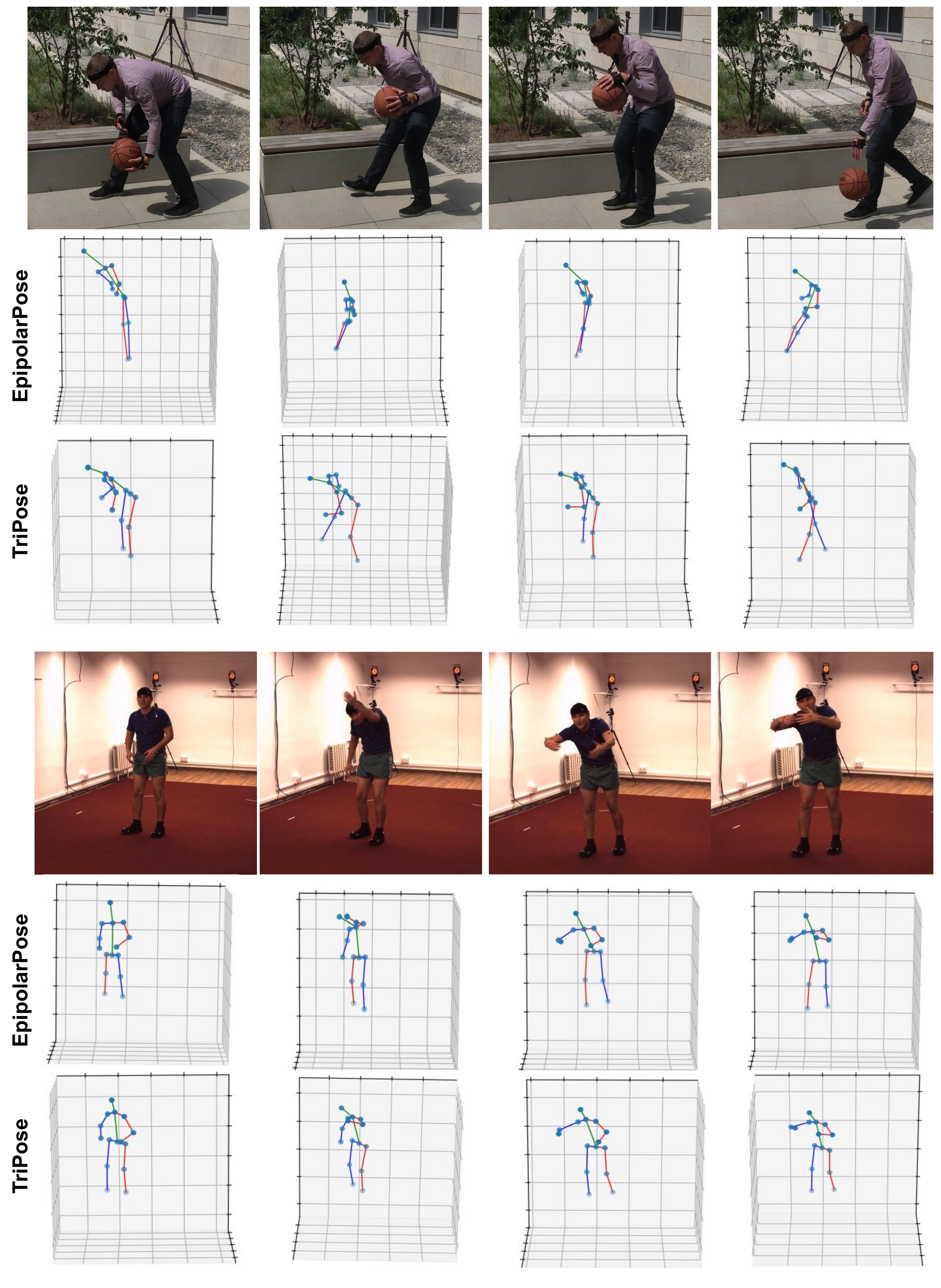}
\end{center}
  \caption{Qualitative comparisons of EpipolarPose \cite{Kocabas_2019_CVPR} and the proposed TriPose using in-the-wild frames from 3DPW dataset and in-door videos from Human3.6M. TriPose and EpipolarPose show promising results in Human3.6 frames; while TriPose outperforms EpipolarPose when testing on unseen frames from 3DPW. }
\label{fig:EpipolarvsTripose}
\end{figure}

\begin{figure*}[t]
\begin{center}
% \fbox{\rule{0pt}{2in} \rule{.6\linewidth}{0pt}}
\includegraphics[scale=0.32]{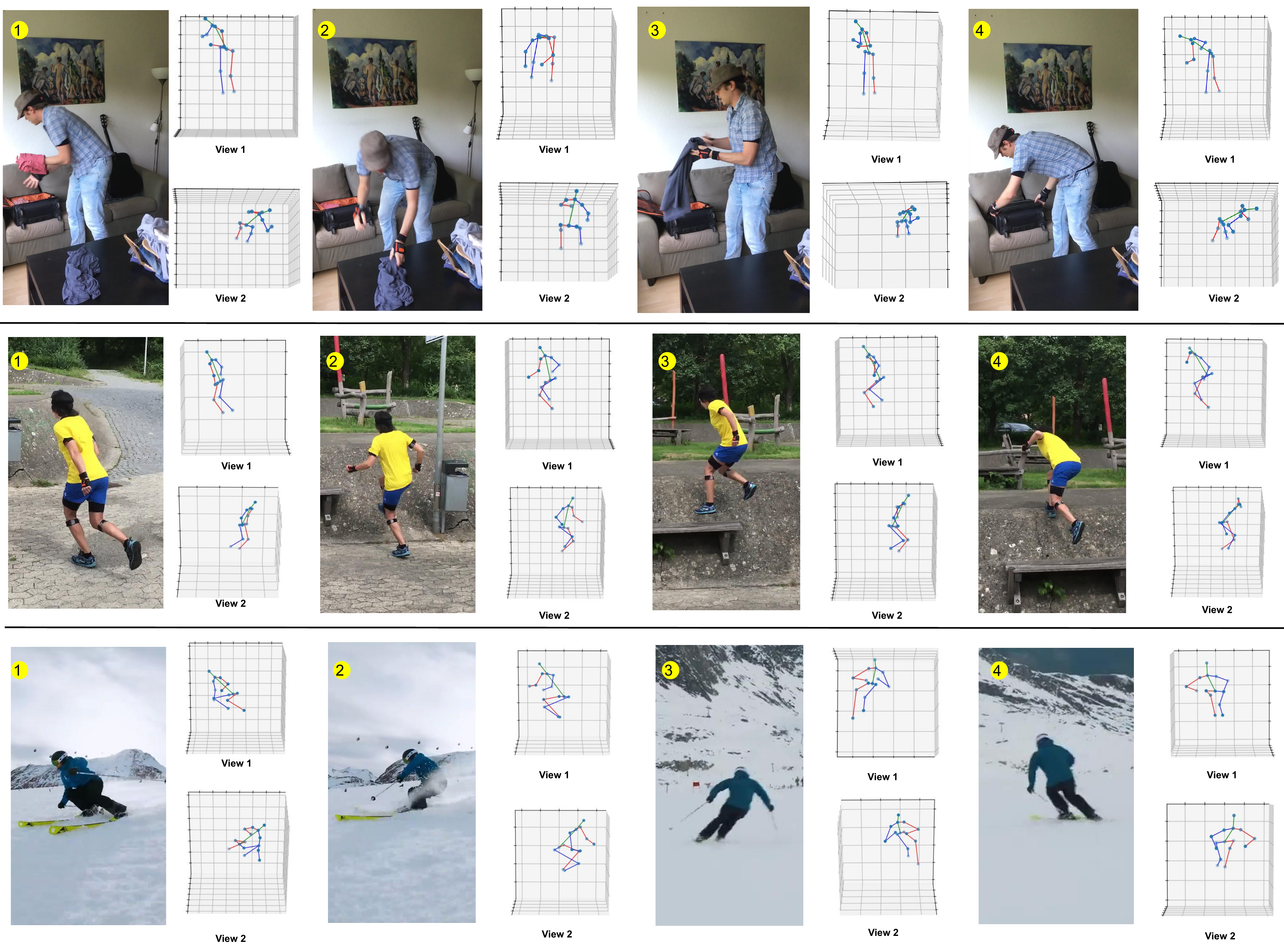}
\end{center}
  \caption{Qualitative analysis examples on 3DPW dataset videos (first and second row) and challenging videos from YouTube (third row).}
\label{fig:YouTube}
\end{figure*}

 We also perform qualitative analysis on 3DPW, which is an in-the-wild dataset. This dataset is not multi-view, therefore we do not fine-tune the model on this dataset. Fig. \ref{fig:YouTube} shows performances of our method on challenging tasks such as ski and running outdoors. Here we use an pretrained AlphaPose for 2D joint detection and bounding boxes are detected using YOLOv3 \cite{redmon2018yolov3}. Fig. \ref{fig:YouTube} shows that temporal information enables the network to track the motion and correctly predict body parts in some of the frames that human joints are occluded. Fig. \ref{fig:EpipolarvsTripose} compares performances of EpipolarPose and TriPose on the test set of Human3.6M and in-the-wild videos from 3DPW. Although EpipolarPose shows promising estimation using Human3.6M images, its predictions for in-the-wild frames are poor. EpipolarPose uses the mean and standard deviation of Human3.6M dataset to normalize the input frames. Therefore, the trained models are not generalizable to unseen datasets. %Figure \ref{fig:YouTube} shows extensive evaluations on two videos from 3DPW dataset and a ski video from YouTube.  

\subsection{Temporal Networks and Self-Supervision}
We evaluate performance of our temporal lifting network on three training schemes including unpaired data annotations using adversarial training, \textit{UP-3D}, multi-view supervision from re-projection only, \textit{MV-Reproj}, and multi-view supervision from triangulation, \textit{MV-Triang}. Particularly, we compare performance of our lifting network with dilated convolutional neural network (dilated-CNN) that was introduced in \cite{Pavllo_2019_CVPR}. Dilated-CNN showed promising performance when strong supervision is provided. Table \ref{tab:Ablation_Temporal} shows that our model outperforms in \textit{UP-3D} and \textit{MV-Reproj}. In \textit{MV-Triang}, the generated pseudo ground truth annotations make the training close to supervised training. Therefore, dilated-CNN outperforms our lifting network by 1.5 mm. We hypothesize that dilated-CNN needs stronger supervision than our lifting network. Our recurrent lifting network outperforms the recurrent lifting network proposed in \cite{Hossain_2018_ECCV} and \cite{Kocabas_2020_CVPR}. \cite{Hossain_2018_ECCV} uses an LSTM-based sequence to sequence model to exploit temporal information. For fully supervised training, Table \ref{tab:Human3.6M} shows that our model obtained an error of 55.4 mm while Hossain et al. \cite{Hossain_2018_ECCV} obtain 58.3 mm error. We observed that using sequence to one output is better than sequence to sequence model. Weakly-supervised adversarial models have been widely used in literature for single frames\cite{Wandt_2019_CVPR, Kanazawa_2018_CVPR}. The proposed temporal network can therefore further improve previous weakly-supervised training frameworks.  

{\renewcommand{\arraystretch}{1}
\begin{table}[t]
\footnotesize
\centering
\caption{Human3.6M dataset: Evaluation of temporal lifting networks for different weakly-supervised training methods.}
\label{tab:Ablation_Temporal}
\begin{tabular}{ p{3cm}|cccc}
\hline
Lifting Net & Supervision & MPJPE  \\
\hline
Pavllo (27 frames) in \cite{Pavllo_2019_CVPR} & UP-3D & 85  \\
GRUA (ours)  & UP-3D & 79 &  \\
Pavllo (27 frames) in \cite{Pavllo_2019_CVPR} & MV-Reproj & 219.8  \\
GRUA (ours) & MV-Reproj & 144.6 \\
Pavllo (27 frames) in \cite{Pavllo_2019_CVPR} & MV-Triang & 62.9 \\
GRUA (ours) & MV-Triang & 64.4 \\
\hline
\end{tabular}
\end{table}}

{\renewcommand{\arraystretch}{1}
\begin{table}[h!]
\footnotesize
\centering
\caption{Human3.6M dataset: Ablation study on supervision elements of the proposed model.}
\label{tab:Ablation_model}
\begin{tabular}{ c|cc|cc}
\hline
 & \multicolumn{2}{c|}{Supervision}  & \multicolumn{2}{c}{} \\
Temp & Triang & Reproj  & MPJPE & PMPJPE \\
\hline
 \ &\checkmark &  & 74.4 & 56.1 \\
 & \checkmark & \checkmark & 72.5 & 54.7\\
 \checkmark & \checkmark&  & 66.0 & 48.8\\
\checkmark & \checkmark & \checkmark & 64.4 & 48.1\\
\hline
\end{tabular}
\end{table}}

\subsection{Ablation Studies}
We perform ablation studies on supervision elements of our framework and on temporal information. Table \ref{tab:Ablation_model} indicates that using temporal information significantly improves our MPJPE results. We use 27 frames as the input. Increasing the number of frames to 81 did not change our results. The idea of using multi-view re-projection along with triangulation could improve the results by 2 mm both in single-frame input and video-input. Triangulation alone with single-frame input obtains an error of 74.4 mm. Kocabas et al. \cite{Kocabas_2019_CVPR} reported the error of 77.0 mm with only triangulation. This validates that the method with only triangulation is a baseline for our work. Our error reduces from 74.4 to 64.4 when we use temporal information and multi-view re-projection.
We also perform an ablation study on the error of 2D keypoints. Table \ref{tab:Ablation_2D} shows the affect of the accuracy of 2D keypoints on the final estimation. The accuracy of 2D keypoints is critical for the lifting network as well as triangulation. Our experiments show that estimation of the rotation matrix by SVD particularly is affected by 2D errors. Using the fine-tuned model or a pretrained model with 8.9 pixel error \cite{Iqbal_2020_CVPR} could obtain an error of less than 40 mm for triangulation. This error level is comparable to that of state-of-the-art fully supervised training models \cite{Chen_2019_CVPR}.       
{\renewcommand{\arraystretch}{1}
\begin{table}[b]
\footnotesize
\centering
\caption{Human3.6M dataset: Ablation study on 2D errors. The unit of 2D error is pixel (px) and units of 3D errors are mm. 3D errors have been reported after Procrustes alignments (P).}
\label{tab:Ablation_2D}
\begin{tabular}{ p{3cm} |c c c}
\hline
Method & 2D (px) & Triang (P) & 3D (P)  \\
\hline
Pretrained (from \cite{Martinez_2017_ICCV})
& 12.5 &  53.6 & 58.5  \\
% AP (PT)\cite{wandt2020canonpose}
% & 11 & - &\\
Pretrained (ours)
& 8.9 & 30.0 &48.8 \\
Fine-tuned (from \cite{Pavllo_2019_CVPR})
& 7.3  & 26.69 & 43.8  \\ \hline
% GT
% &  0  & 18.77 &\\
% GT + $\mathcal{N}(0,5)$
% &  6.3  & 45.34 & 51.2\\
% GT +  $\mathcal{N}(0,10)$
% % &  12.53  & 78.85 & 56.1\\
% GT + $\mathcal{N}(0,15)$
% &  18.79  & 112.162 & 71.5\\
%\hline
\end{tabular}
\end{table}}
\subsection{Conclusions and Future Work}
Weakly-supervised training for 3D pose estimation has been attracting increasing attention \cite{Wandt_2019_CVPR, Iqbal_2020_CVPR, Kocabas_2019_CVPR}. However, previous works did not exploit temporal information for weakly-supervised training. In this work, we showed that designing a temporal encoder for weakly-supervised training is challenging, as temporal models introduced for fully supervised training are not necessarily suitable for weakly-supervised setting. Furthermore, we showed that triangulation from multi-views together with multi-view re-projection outperforms previous multi-view self-supervision methods. We achieved state-of-the-art performance on benchmark datasets. In the future works, we plan to improve the proposed method from several aspects. Currently we freeze the triangulation part and the error of the multi-view re-projection does not back-propagate through SVD. Therefore, each view contributes equally to the triangulation. Future work should address this by implementing a learnable triangulation\cite{Iskakov_2019_ICCV}. Our experiments showed that the error of 2D poses is critical in triangulation. However, our networks does not back-propagate the view-consistency losses to improve 2D poses. Future works should implement an end-to-end framework that accepts sequence of frames as its input.

% {\renewcommand{\arraystretch}{1}
% \begin{table}[h!]
% \footnotesize
% \centering
% \caption{Evaluation results of the trained network for the Skii dataset in Protocol I and II. All numbers are in millimeters and the lower the better.}
% \label{tab:protocol I}
% \begin{tabular}{ p{2cm} p{1.3cm} c{0.6cm}c{0.7cm}c{0.7cm} }

% \hline
% Method & Supervision &  MPJPE & NMPJPE & PMPJPE \\
% \hline
% Rhodin \cite{Kocabas_2019_CVPR}
% & MV & - &   &   \\
% Wandt \cite{wandt2020canonpose}
% & MV &  & - &   \\
% Ours 
% & MV &  \textbf{} & \textbf{} & \textbf{} \\ 
% \hline
% \end{tabular}
% \end{table}}

% {\renewcommand{\arraystretch}{1}
% \begin{table}[b!]
% \footnotesize
% \centering
% \label{tab:Generator}
% \caption{Comparison of different generator networks}
% \begin{tabular}{c|cc}
% \hline
% Generator & Protocol I (mm) & Protocol II (mm) \\
% \hline
% \rowcolor{Gray}
% CNN \cite{Pavllo_2019_CVPR} & 52.4 &  37.6\\
% FC-GRU & 43.7 & 31.6\\
% \hline
% \end{tabular}
% \end{table}}
%------------------------------------------------------------------------

{\small
\bibliographystyle{ieee_fullname}
\bibliography{egbib}
}

\end{document}